%% file: main.tex
\documentclass[10pt,twocolumn,letterpaper]{article}

\usepackage[pagenumbers]{wacv} 

\input{preamble}

\input{commands}
\usepackage{dsfont}

\definecolor{wacvblue}{rgb}{0.21,0.49,0.74}
\usepackage[pagebackref,breaklinks,colorlinks,allcolors=wacvblue]{hyperref}
\usepackage{booktabs,multirow}
\usepackage{xcolor}

\def\confName{WACV}
\def\confYear{2026}

\title{HumanSplatHMR: Closing the Loop Between Human Mesh Recovery and Gaussian Splatting Avatar}

\author{
Yeheng Zong$^*$ \quad Pou-Chun Kung\thanks{Denotes equal contribution.} \quad Yike Pan \quad Seth Isaacson \quad Yizhou Chen \\ Ram Vasudevan \quad Katherine A. Skinner \\
University of Michigan \\
{\tt\small \{yehengz, pckung, yikepan, sethgi, yizhouch, ramv, kskin\}@umich.edu}
}

\begin{document}
\maketitle

\begin{abstract}
Accurately recovering human pose and appearance from video is an essential component of scene reconstruction, with applications to motion capture, motion prediction, virtual reality, and digital twinning.
Despite significant interest in building realistic human avatars from video, this paper demonstrates that existing methods do not accurately recover the 3D geometry of humans.
ViT-based approaches are not consistently reliable and can overfit to 2D views, while NeRF- and Gaussian Splatting–based avatars treat pose and appearance separately, limiting rendering generalization to new poses.
To resolve these shortcomings, this paper proposes \textbf{HumanSplatHMR}, a joint optimization framework that refines 3D human poses while simultaneously learning a high-fidelity avatar for novel-view and novel-pose synthesis. Our key insight is to close the loop between geometric pose estimation and differentiable rendering.
Unlike prior human avatar methods that rely on accurate human pose obtained through motion capture systems or offline refinement, which are impractical in in-the-wild scenarios, our approach uses only human mesh estimates from a state-of-the-art human pose estimator to better reflect real-world conditions.
Therefore, instead of using the human pose only as a deformation prior, HumanSplatHMR backpropagates photometric, segmentation, and depth losses through a differentiable renderer to the pose parameters and global position.
This coupling refines the global 3D pose over time, improving accuracy and alignment while producing better renderings from novel views.
Experiments show consistent improvements over pose recovery baselines that omit image-level refinement and avatar baselines that decouple pose estimation from avatar reconstruction. 
A project page is available at \href{https://scottyehengz.github.io/HumanSplat}{https://scottyehengz.github.io/HumanSplat}.


\end{abstract}

\begin{figure}[t!]
    \centering
    \includegraphics[width=1.0\linewidth]{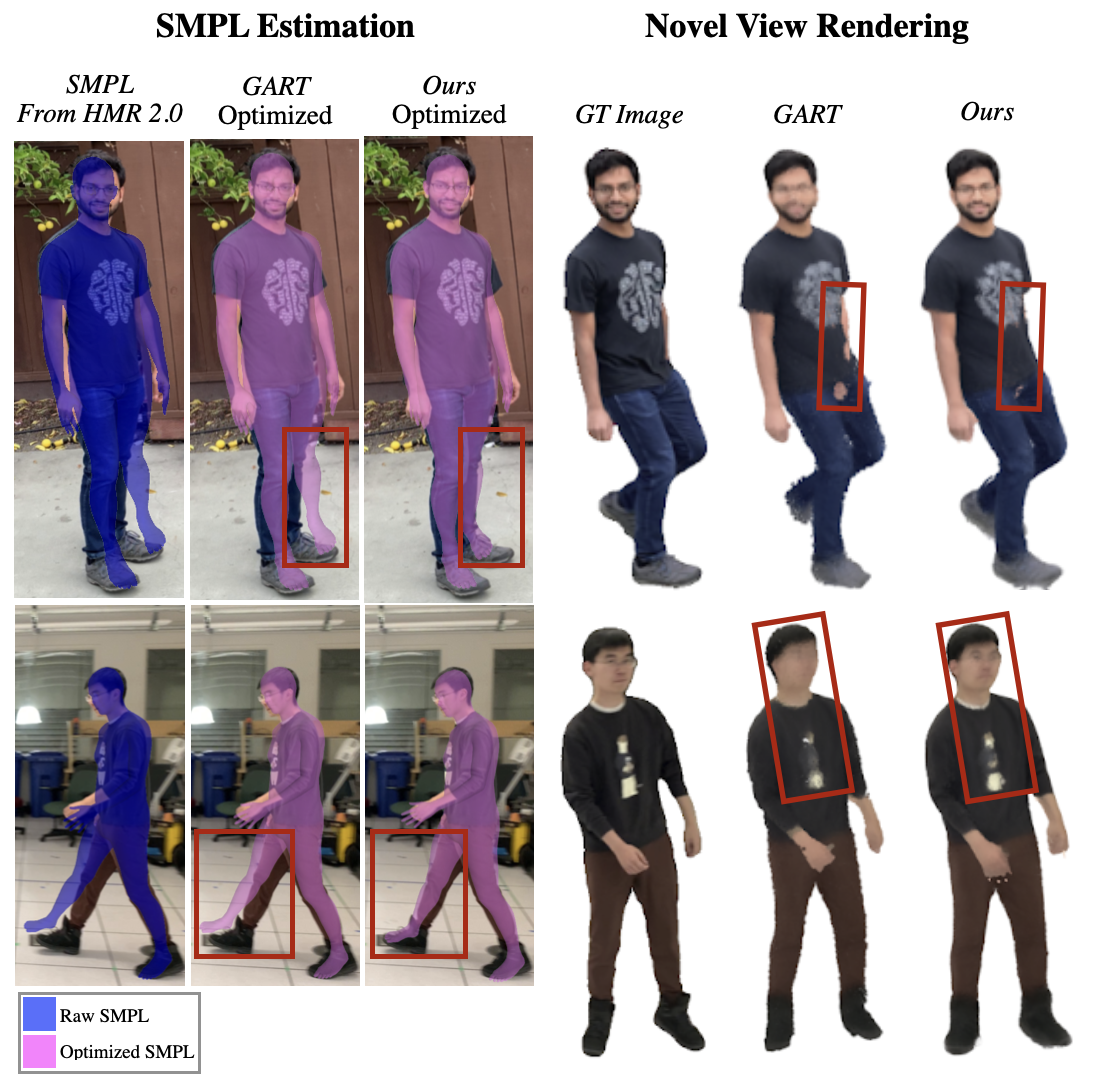}
        \caption{
        HumanSplatHMR advances both SMPL estimation and novel-view rendering by enabling joint optimization of the human mesh and Gaussian splats, taking SMPL estimates from HMR 2.0~\cite{4dhuman} as input.
        In contrast, prior works, such as GART~\cite{gart}, rely on accurate SMPL from motion capture or refined SMPL to deform Gaussians for avatar reconstruction.
        These methods also deliberately decouple SMPL from Gaussian splats to improve rendering quality, but this prevents refinement of human pose estimation and ultimately results in sub-optimal novel-view performance.
        HumanSplatHMR addresses this limitation and allows human pose and Gaussian splats to mutually refine each other. As a result, it achieves more accurate SMPL estimation and consistently higher rendering quality.
        }
        \label{fig:teaser_fig}
\end{figure}

\section{Introduction}
\label{sec:intro}




Photorealistic human rendering and human mesh reconstruction are two fundamental and complementary goals in computer vision and graphics, with wide-ranging applications in AR/VR, gaming, simulation, virtual try-on, and cinematic production~\cite{film}. While photorealistic rendering enables immersive visual experiences, mesh reconstruction provides explicit and animatable geometry that supports editing, motion retargeting, physics-based simulation, and physical interaction. 
Despite their importance, reconstructing realistic avatars and accurate human meshes from monocular video remains highly challenging. Monocular human mesh reconstruction methods often struggle to produce meshes that align well with the image.
Photorealistic avatar approaches typically use human mesh reconstruction as a motion prior for novel-view and novel-pose rendering. However, these methods generally overlook errors in mesh reconstruction and over-rely on the estimated meshes, which can lead to suboptimal avatar quality and degraded realism.

Recent studies of Human Mesh Recovery (HMR) center on the Skinned Multi-Person Linear (SMPL)~\cite{smpl} model, which provides a parametric representation of the human body with a compact set of shape parameters \cite{hmr}. 
Recent end-to-end HMR methods directly regress SMPL parameters either from a single image~\cite{hmr,4dhuman,tokenhmr} or from video sequences with temporal cues~\cite{vibe}, achieving promising results but still prone to imperfect alignment and overfitting to 2D images. 
Based on SMPL estimation methods, human avatar modeling has advanced through the application of radiance field rendering models, such as neural radiance fields (NeRF) \cite{nerf} or 3D Gaussian Splatting (3DGS \cite{3dgs}). However, there exists a trade-off in existing methods between accurate SMPL estimation and high fidelity rendering quality.
Methods that optimize SMPL with photorealistic representation tightly-coupled to SMPL achieve accurate human geometry, but sacrifice rendering fidelity~\cite{ihuman}. Conversely, approaches that allow photorealistic representation to detach from SMPL through learnable skinning weights or deformation field deliver high-quality renderings, yet degrade SMPL accuracy~\cite{hugs, gart, gauhuman, splattingavatar}. This trade-off between geometry and appearance remains unresolved.

In this work, we propose HumanSplatHMR, a unified framework that bridges these gaps, as shown in Figure~\ref{fig:teaser_fig}. Firstly, our method combines end-to-end HMR with a rendering-based iterative refinement stage.
In particular, HumanSplatHMR leverages Gaussian representations supervised by photometric, depth, and segmentation cues instead of 2D keypoints.
As a result, experiments demonstrate that HumanSplatHMR supports global pose estimation, reduces overfitting to 2D views, and improves alignment of the estimated mesh and pose to the ground-truth.
Further, to mitigate the trade-off between geometry and appearance in human avatar modeling, we introduce a joint optimization between the Gaussian representation and human poses with simple-yet-effective Cloth-Aware Mesh-Embedded Loss (CAMEL) that improves both human pose estimation and novel view rendering. Together, our method yields consistent improvements in both SMPL accuracy and photorealistic rendering, enabling the reconstruction of high-fidelity human avatars from video.


\section{Related Work}

\subsection{Human Mesh Recovery}
Human mesh recovery is commonly performed using the SMPL model~\cite{smpl}, whose parameters can represent different human poses and shapes. Many early approaches estimated 3D human pose and shape through iterative optimization. The first fully automatic method, SMPLify~\cite{SMPLify}, proposed to instead fit SMPL to 2D keypoint detections obtained from an off-the-shelf keypoint detector, guided by strong priors to stabilize optimization. However, this method is slow and only uses sparse supervision at the keypoint locations.

Subsequent research introduced direct prediction of SMPL parameters using pre-trained models. Kanazawa et. al.~\cite{hmr} introduced the first end-to-end learning-based model for human pose recovery. SPIN~\cite{spin} extended this idea by incorporating an “in-the-loop” optimization that embeds SMPLify within HMR training using a CNN-based model. 
To further address temporal consistency, many approaches add sequential modeling mechanisms, such as a temporal encoder that fuses per-frame features~\cite{hmmr} or recurrent temporal encoders to smooth pose predictions~\cite{vibe, meva}. Beyond single-image human recovery, PHALP~\cite{phalp} extends HMR frameworks to human tracking, exploiting 3D reconstruction for robust identity association over time. Still, the performance is limited by single-frame estimation. 

Most recently, HMR approaches have been extended to ViT-based methods such as HMR2.0~\cite{4dhuman}, TokenHMR~\cite{tokenhmr}, and HSMR~\cite{HSMR}, which have shown promising performance by improving speed and accuracy. However, end-to-end HMR methods often suffer from sub-optimal 3D alignment and overfitting to 2D views. 

\begin{figure*}[t]
    \centering
    \includegraphics[width=0.99\linewidth]{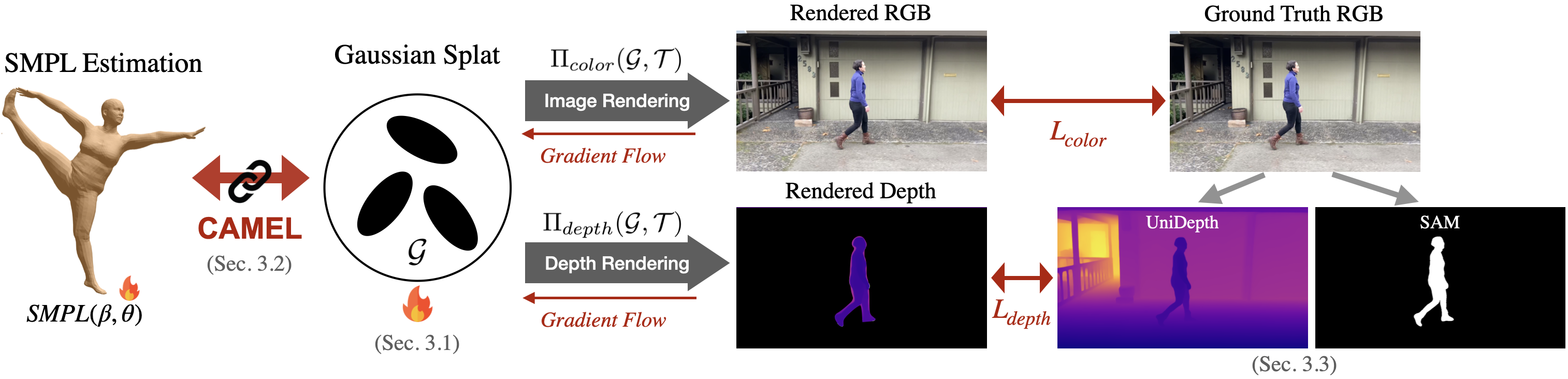}
        \caption{
        Method Overview. HumanSplatHMR takes SMPL estimation as input and combines SMPL with Gaussians with the proposed CAMEL. This enables both SMPL and Gaussian representation optimization using rendered color and depth images.
        }
        \label{fig:overview}
\end{figure*}

\begin{figure}[b!]
    \centering
    \includegraphics[width=0.99\linewidth]{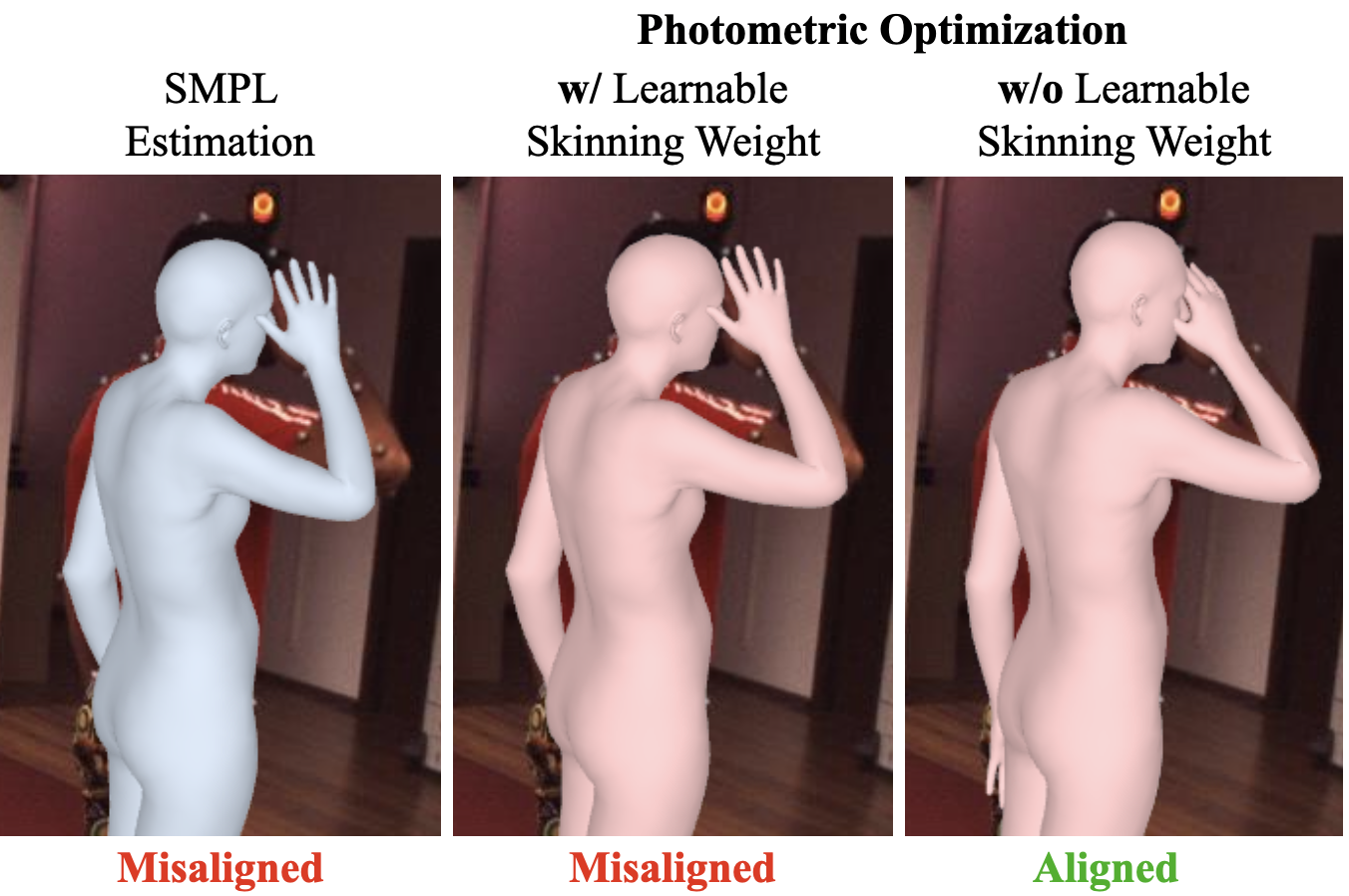}
        \caption{
        Raw SMPL estimation and photometric optimized SMPL with and without learnable skinning weights. Enabling learnable deformation like learnable skinning weights allows Gaussians to detach from the SMPL mesh, which causes sub-optimal SMPL refinement.
        }
        \label{fig:learnabe skinning weight}
\end{figure}

\subsection{Human Avatar}
Recent advances in NeRF~\cite{nerf} and Gaussian Splatting~\cite{3dgs} have significantly accelerated progress in human avatar modeling, enabling photorealistic rendering of humans.
These renderings can be formed both from novel camera views and novel human joint configurations. NeRF-based human avatar methods typically leverage a skeleton or the SMPL model to construct a deformation field, allowing reconstruction from monocular video and articulation of the human body after training~\cite{neuralbody, Anim-NeRF, humannerf, instantavatar}.

Gaussian Splatting–based avatar methods~\cite{hugs, gart, gauhuman, splattingavatar, gomavatar} adopt a similar idea by deforming Gaussians according to the SMPL mesh using classical linear blend skinning (LBS)~\cite{lbs}. 
However, relying exclusively on fixed SMPL estimates to deform Gaussians propagates errors and leads to degraded performance when SMPL predictions fail, while also limiting the ability to capture non-rigid clothing.
To address this limitation, recent human Gaussian Splatting works enable joint optimization of SMPL and Gaussian splats by incorporating a rendering-based photometric loss. Most of the works demonstrated improved rendering quality by using learnable skinning weights~\cite{hugs, gart, gauhuman, splattingavatar} or non-rigid deformations~\cite{gomavatar}. We compare against GART~\cite{gart} as the primary baseline in our evaluations, chosen as a representative method that performs well.
However, this added flexibility reduces SMPL accuracy: once the Gaussians are allowed to decouple from the mesh, they are no longer constrained to follow SMPL geometry.
As a result, the rendering may overfit to training views while the SMPL pose remains wrong, as shown in Figure~\ref{fig:learnabe skinning weight}.
While this leads to strong evaluations when the test set remains close to the training set, the models do not generalize to views or poses too different from those seen during training.
Conversely, iHuman~\cite{ihuman} tightly couples Gaussians with the SMPL mesh using mesh-embedded Gaussians. This enables SMPL refinement and improves pose estimation, but the strict coupling reduces flexibility, leading to worse novel view rendering.

In HumanSplatHMR, we introduce a Cloth-Aware Mesh-Embedded Loss (CAMEL) that establishes a loose but consistent coupling between the Gaussians and the SMPL model. This formulation simultaneously guides Gaussian placement with mesh-aware priors and allows local flexibility for clothing and non-rigid deformations. As a result, CAMEL not only facilitates accurate refinement of SMPL parameters but also enhances rendering fidelity, yielding superior novel-view synthesis and robust human pose generation.





\section{Method}
In this section, we first review the preliminaries and existing approaches that combine Gaussian Splatting with the SMPL model. We then introduce a novel loss function that loosely couples Gaussians with the SMPL mesh. Finally, we present the full loss function used for optimization. The method overview is shown in Figure~\ref{fig:overview}.

\subsection{Preliminaries}
\subsubsection{Gaussian Splatting}
3D Gaussian Splatting~\cite{3dgs} models the 3D scene using a set of Gaussian functions. Each 3D Gaussian is parameterized by its position $\mu$, rotation quaternion $q$, scaling vector $S$, opacity $\alpha$, and spherical harmonic (SH) coefficients $sh$.
\begin{equation}
    \mathcal{G} = \{G_i := (\mu_i, q_i, S_i, \alpha_i, {sh}_i)\}_{i=1}^N
\end{equation}
The covariance of each Gaussian is computed according to
\begin{equation}
    \Sigma = RSS^TR^T,
\end{equation}
where $S \in \mathbb{R}^3$ is a 3D scale vector with the singular values of $\Sigma$, and $R \in \text{SO}(3)$ is rotation matrix computed from quaternion $q$.
The color and depth image can be rendered from camera view $\mathcal{T}$ following~\cite{3dgs}. We denote the rendering color and depth rendering equation as $\hat{\mathbf{I}} = \Pi_{color}(\mathcal{G}, \mathcal{T})$ and $\hat{\mathbf{D}}= \Pi_{depth}(\mathcal{G}, \mathcal{T})$, respectively.

\begin{figure*}[h!]
    \centering
    \includegraphics[width=0.90\linewidth]{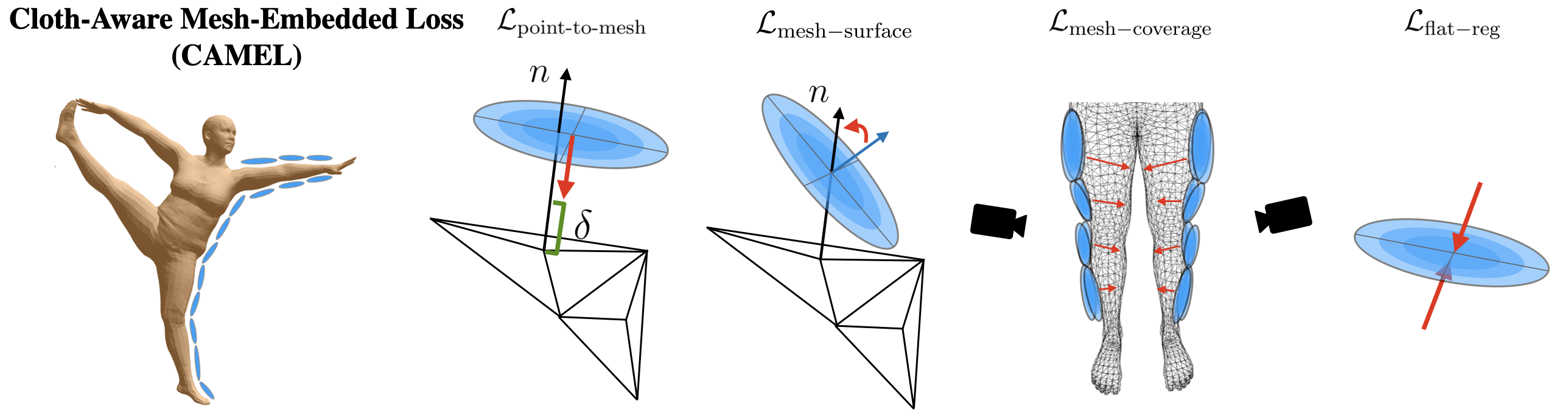}
        \caption{
        Cloth-Aware Mesh-Embedded Loss (CAMEL) illustration. CAMEL loosely couples the SMPL mesh with the Gaussian representation. The key motivation is to better model clothing by allowing local non-rigid deformations. CAMEL constrains Gaussians to remain close to the human mesh while enforcing surface alignment and ensuring full mesh coverage. $n$ is the normal of the mesh vertex and $\delta$ represents the tolerance margin between the cloth and the body.
        }
        \label{fig:camel illustration}
\end{figure*}

\subsubsection{SMPL Model and Template}

The SMPL model~\cite{smpl} provides a compact parametric representation of the human body.
It takes as input pose parameters $\boldsymbol{\theta} \in \mathbb{R}^{24 \times 3 \times 3}$ and shape parameters $\boldsymbol{\beta} \in \mathbb{R}^{10}$,
and generates a mesh with vertices $\mathbf{V} \in \mathbb{R}^{3 \times N}$, where $N = 6890$.
The pose vector $\boldsymbol{\theta}$ is composed of the body pose $\boldsymbol{\theta}_b \in \mathbb{R}^{23 \times 3 \times 3}$ together with a global orientation term $\boldsymbol{\theta}_g \in \mathbb{R}^{3 \times 3}$. The vertices of an SMPL mesh in canonical space can be obtained by $\mathbf{V}_c = \mathbf{T}(\boldsymbol{\beta})$, where $\mathbf{T}$ is the shape template.
The mesh can be deformed using linear blend skinning (LBS) based on articulated pose $\boldsymbol{\theta}$:
\begin{equation}
    \mathbf{V}(\boldsymbol{\beta}, \boldsymbol{\theta}) 
    = \left( \sum_{k=1}^{n_b} w_k(\mathbf{V}_c(\boldsymbol{\beta})) \, B_k(\boldsymbol{\theta}) \right) \mathbf{V}_c(\boldsymbol{\beta}),
\end{equation}
where $B$ is the rigid body bones based on pose $\boldsymbol{\theta}$, $n_b$ is the number of bones, and $w_k(x)$ is the skinning weight describing how each vertex is influenced by bones. Note that skinning weight is precomputed and fixed for the SMPL template mesh.

In this work, we leverage detected SMPL models $(\boldsymbol{\beta}, \boldsymbol{\theta})$ from HMR2.0~\cite{4dhuman}, then optimize pose parameters $\boldsymbol{\theta}$ through 
joint optimization of human pose and Gaussian parameters.

\subsubsection{Human Gaussian Splat Representation}
To combine Gaussian splatting with the SMPL model, Gaussians can also be deformed based on pose $\boldsymbol{\theta}$ using LBS:
\begin{equation}
    \boldsymbol{\mu}'(\boldsymbol{\beta}, \boldsymbol{\theta}) 
    = \left( \sum_{k=1}^{n_b} w_k (\boldsymbol{\mu}) \, B_k(\boldsymbol{\theta}) \right) \boldsymbol{\mu},
\end{equation}
where $\boldsymbol{\mu}'$ denotes Gaussian center deformed from $\boldsymbol{\mu}$.
To enable more photorealistic rendering and non-rigid cloth-modeling, most prior Gaussian avatar works~\cite{hugs, gart, gauhuman, splattingavatar} propose learnable skinning $\Delta \mathbf{w}$ to allow non-rigid motion:
\begin{equation}
\widehat{{w}}(\boldsymbol{\mu}) = {w}(\boldsymbol{\mu}) + \Delta \mathbf{w},
\end{equation}
or apply learnable non-rigid deformation fields~\cite{gomavatar}.

However, the LBS skinning weight of an SMPL template is pretrained using high-quality and expensive 3D human model scans. Re-enabling skinning weight optimization can improve rendering fidelity by letting the projected mesh overfit the 2D image, but it hurts both novel view or pose rendering and SMPL pose estimation. First, the learnable skinning allows Gaussians to detach from the human mesh, which makes the Gaussian splat overfit to 2D training views. This can severely degrade rendering quality when rendering from views far from training views. Secondly, due to the disconnect between SMPL meshes and Gaussian splats, the mesh refinement can degrade SMPL estimation, as shown in Figure~\ref{fig:teaser_fig}.


\subsection{Cloth-Aware Mesh-Embedded Loss}
\label{sec:camel}
To achieve high-fidelity rendering without the overfitting and mesh-disentanglement induced by learnable skinning weights, we propose the Cloth-Aware Mesh-Embedded Loss (CAMEL). CAMEL enables high-quality avatar generation with 3D Gaussian Splatting while maintaining the geometric correlation between Gaussians and the SMPL model.
The loss is designed to keep Gaussians close to the human mesh within a cloth margin, ensure full coverage of the human body, and encourage alignment of Gaussians with the SMPL surface.
As a result, CAMEL ensures both geometric and rendering consistency. The illustration is shown in Figure~\ref{fig:camel illustration}.

\vspace{-0.2in}


\paragraph{Cloth-Aware Point-to-Mesh Surface Loss.}
Rather than forcing the Gaussians to stay strictly on the mesh surface, we propose a point-to-plane loss that keeps Gaussians within a thin band of thickness $\delta$ around the SMPL mesh.
In particular, we compute this loss as
\begin{align}
\boldsymbol{d}_{p2mesh} &= \langle \boldsymbol{\mu}_i - \mathbf{v}_{\mathrm{NN}(\boldsymbol{\mu}_i)}, \,\mathbf{n}_{\mathrm{NN}(\boldsymbol{\mu}_i)} \rangle \\
\mathcal{L}_{\text{p2mesh}} &= \frac{1}{N} \sum_{i=1}^N 
\max \!\Big( \,\big| \boldsymbol{d}_{p2mesh} \big| - \delta, \; 0 \,\Big)
\end{align}
where $\mathbf{n}$ denotes the surface normal of the SMPL vertices and $\mathbf{v}_{NN(\boldsymbol{\mu}_i)}$ and $\mathbf{n}_{NN(\boldsymbol{\mu}_i)}$ denotes the position and normal of nearest neighbor SMPL vertex of $\boldsymbol{\mu}_i$. 
\paragraph{Mesh Coverage Loss.} 
To ensure full-body Gaussian coverage, we require each mesh vertex $\mathbf{v}$ to be close to at least one Gaussian:
\begin{equation}
\mathcal{L}_{\mathrm{coverage}}=\mathbb{E}_{\mathbf{v}}\!\left[\max\bigl(\lVert \mathbf{v}-\boldsymbol{\mu}_{\mathrm{NN}(\mathbf{v})}\rVert-\delta,\,0\bigr)\right].
\end{equation}


\paragraph{Gaussian Flatness Regularization.} 
Inspired by recent works~\cite{autosplat, splatface, sadgs, sugar, 2dgs} demonstrating that flattened, surface-aligned Gaussians improve novel-view rendering quality, we introduce a regularization loss that encourages Gaussian shapes to remain flat and aligned with the mesh surface. 
For each Gaussian scale 
$\mathbf{S}_i = [s_{0}, s_{1}, s_{2}]$ with $s_0 \leq s_1 \leq s_2$, 
we take $s_n = s_0$ as the normal-axis scale and 
$\{s_{t1}, s_{t2}\} = \{s_1, s_2\}$ as the tangential-axis scales. 
We encourage the tangential axes to exceed the normal axis by a ratio $\tau$:  

\begin{equation}
    \mathcal{L}_{\mathrm{flatness}} 
    = \bigl|\log \tfrac{s_{t1}}{s_n} - \log \tau \bigr| 
    + \bigl|\log \tfrac{s_{t2}}{s_n} - \log \tau \bigr|.
\end{equation}



\paragraph{Surface Alignment Loss.} Complementary to the flatness regularization, we add a loss encouraging the smallest axis of the Gaussian to align with the SMPL surface normal:
\begin{equation}
\mathcal{L}_{\mathrm{surface}}= \frac{1}{N} \sum_{i=1}^N (1-\bigl|\langle \mathbf{R}_i\mathbf{e}_n, \mathbf{n}_{NN(\boldsymbol{\mu}_i)} \rangle\bigr|),
\end{equation}
where $\mathbf{e}_n$ denotes the canonical axis associated with $s_n$ and $\mathbf{R}$ is the rotation matrix of a Gaussian.

The total CAMEL loss is:

\begin{equation}
\mathcal{L}_{\mathrm{CAMEL}} = \lambda_{1} \mathcal{L}_{\mathrm{p2mesh}} + \lambda_{2} 
 \mathcal{L}_{\mathrm{coverage}} + \lambda_{3} \mathcal{L}_{\mathrm{flatness}} + \lambda_{4} \mathcal{L}_{\mathrm{suface}}, 
\end{equation}
where $\lambda_i$ denotes the weights associated with the respective loss terms.

\subsection{Color and Depth Supervision}
In addition to the geometric losses proposed in Section~\ref{sec:camel}, rendering-based losses are applied to directly supervise the rendering process.

\subsubsection{Photometric Loss}
Given the rendered image $\hat{\mathbf{I}}$ and the ground-truth frame $\mathbf{I}$, the L1 loss $\mathcal{L}_{L1}=|\hat{\mathbf{I}}-\mathbf{I}|$ and SSIM loss $\mathcal{L}_{SSIM}$ are    computed using the image similarity metric~\cite{ssim}. We then minimize photometric error using:
\begin{equation}
\mathcal{L}_{color} = \mathcal{L}_{L1} + \lambda_{SSIM} \mathcal{L}_{SSIM}
\end{equation}

\subsubsection{Depth Loss}
Unlike previous works that only produce a SMPL mesh to fit the 2D view without considering global 3D scale and pose, we aim to estimate the SMPL mesh with global scale and pose. We leverage monocular depth estimation output from UniDepthV2~\cite{unidepthv2} to supervise training, thereby improving global 3D pose estimation. 

\paragraph{Rendered Depth Loss.}
Let $\hat{\mathbf{D}}$ be the rendered depth and $\mathbf{D}$ the ground-truth metric depth. We supervise only pixels with positive ground-truth depth values and with a human mask generated by Segment Anything Model 2 (SAMv2)~\cite{samv2}. 

\begin{align}
\mathcal{L}_{\text{depth-2D}} = \left| \mathds{1}_{\{\mathbf{D} > 0, \ \text{SAM} > 0\}} \odot (\hat{\mathbf{D}} - \mathbf{D}) \right|,
\end{align}

\paragraph{Gaussian Center Loss.}
From the current frame, we extract the set of visible Gaussian centers $\mathbf{\mu}$ and construct a depth point cloud $\mathbf{p} = \pi^{-1}(\mathbf{D}, K)$ by inverse projection, where $D$ is the depth image and $K$ is the camera intrinsic. For each Gaussian center $g \in \mathbf{\mu}$, we compute its squared Euclidean distance to the closest point in the depth point cloud. We minimize these nearest-neighbor distances using the Huber penalty~\cite{huber}:
\begin{equation}
\mathcal{L}_{\mathrm{depth-3d}}=
\frac{1}{2|\mathbf{\mu}|}\sum_{\mathbf{g}\in \mathbf{\mu}}\rho_\beta\!\Bigl(\!\min_{p\in \mathbf{p}}\lVert g-p\rVert^2\Bigr),
\label{eq:vtx}
\end{equation}
where $\rho_\beta(u)=\sqrt{u+\epsilon^2}-\epsilon$ and $\epsilon$ is the scale parameter of the Huber penalty and $u$ is Gaussian mean.
This anchors the learned geometry to metric depth while tolerating noise and outliers.

In summary, the depth supervision consists of two components:
\begin{equation}
\mathcal{L}_{depth} = \mathcal{L}_{depth-2D} + \mathcal{L}_{depth-3D}
\end{equation}

\subsubsection{Full Loss}
The full objective combines image and geometry terms:
\begin{equation}
\mathcal{L}
=\mathcal{L}_{color}
+\lambda_{\mathrm{depth}}\mathcal{L}_{\mathrm{depth}}
+\lambda_{\mathrm{CAMEL}}\mathcal{L}_{\mathrm{CAMEL}}
\label{eq:total}
\end{equation}
, where each $\lambda$ denotes the weight associated with the respective loss terms.

\begin{table}[t]
\centering
\small
\setlength{\tabcolsep}{6pt}
\scalebox{0.85}{
\begin{tabular}{l|cc|ccc}
\toprule
\multirow{2}{*}{Method} &
\multicolumn{2}{c|}{H36M} &
\multicolumn{3}{c}{3DPW} \\
\cmidrule(lr){2-3}\cmidrule(lr){4-6}
 & mpjpe$\downarrow$ & \shortstack{pa-mpjpe}$\downarrow$
 & mpjpe$\downarrow$ & \shortstack{pa-mpjpe}$\downarrow$ & v2v $\downarrow$ \\
\midrule
HMR2.0                      &80.58  &\textbf{41.13}  &75.64  &56.26 &239.27  \\
GART          &80.58  &\textbf{41.13}  & 75.64  &56.27 &239.27   \\
Ours          &\textbf{80.53}  &\textbf{41.13}  &\textbf{72.63}  &\textbf{55.10} &\textbf{196.95}  \\
\bottomrule
\end{tabular}
}
\caption{Evaluation of SMPL pose accuracy on H36M and 3DPW datasets. HumanSplatHMR consistently achieves the lowest pose error.}
\label{tab:smpl_pose}
\end{table}

\input{tables/tab3_horizontal}

\input{tables/all_rendering}

\section{Experiments}
This section details the experiments used to evaluate the performance of HumanSplatHMR. 

\subsection{Implementation Details}
We use Pytorch for the implementation of our joint optimization framework that refines the 3D human poses when reconstructing an expressive human avatar.
All tests were run using the Adam optimizer.
We conducted all experiments on a single NVIDIA Tesla V100 SXM2 16GB GPU. We used SMPL as the initial mesh template with its blend skinning weight. We initialized Gaussians on the vertices of the default mesh with $V = 6890$ and $F = 13776$. During training, we take a monocular RGB video as input, accompanied by SMPL parameters, binary mask, metric depth map, and camera parameters at each frame. To best mimic the real-world deployment where most likely only RGB video is available, we use HMR2.0~\cite{4dhuman} for SMPL parameters~\cite{smpl}, SAMv2~\cite{samv2} for binary masks, and UniDepthv2~\cite{unidepthv2} for metric depth maps and camera parameters. Our method runs at $1000$ iterations per second during training with $>120$ fps during inference on NVIDIA V100. During test-time optimization, we follow the design in InstantAvatar~\cite{instantavatar} and GART~\cite{gart} to pass the gradient from Gaussian parameters to optimize SMPL parameters.
\subsection{Dataset}
We evaluate our method on four different datasets, including controlled studio capture and in-the-wild scenes, and select sequences that contain only one human subject.

\noindent\textbf{Human3.6M (H36M)~\cite{h36m}.} H36M is a large-scale indoor motion-capture dataset with calibrated multi-camera studio recordings of multiple subjects performing diverse actions. It is widely used to train human pose and mesh recovery networks for accurate 3D joint keypoint annotations and camera calibration. We use H36M primarily to assess the SMPL pose accuracy.
\par\noindent\textbf{NeuMan~\cite{neuman}.} The NeuMan dataset consists of $6$ videos lasting between $10$ to $20$ seconds and features a single human subject captured using a mobile phone. It offers calibrated sequences with optimized SMPL poses as reference ground truth. We use NeuMan to assess both the rendering quality of novel views and pose synthesis and SMPL pose accuracy.
\par\noindent\textbf{3D Poses in the Wild (3DPW)~\cite{3dpw}.} 3DPW is an in-the-wild dataset with handheld videos and challenging poses, clothing, and lighting. It includes accurate ground-truth 3D poses obtained from wearable sensors, enabling robust evaluation under real-world conditions. We use 3DPW primarily to assess the SMPL pose accuracy.

\subsection{Baselines}
\subsubsection{SMPL Baseline}
We compare against three complementary SMPL-based methods. \textbf{HMR2.0}~\cite{4dhuman} is a fully transformer-based human-mesh-recovery backbone that performs per-image SMPL mesh/pose recovery with strong single-frame accuracy. 

\subsubsection{Human-Avatar Baseline}
We also benchmark against an animatable avatar method tailored for monocular video. 
With a SMPL template prior, \textbf{GART}~\cite{gart} builds Gaussian splatting avatars by jointly optimizing 3D Gaussian representations and SMPL representation with learnable skinning weights enabled as proposed. 

Note that, unlike prior works that rely on accurate SMPL parameters from motion capture systems or offline refinement, we use SMPL estimates from HMR2.0~\cite{4dhuman} as input for all baselines to simulate in-the-wild conditions.

\begin{figure*}[t]
    \centering
    \includegraphics[width=0.99\linewidth]{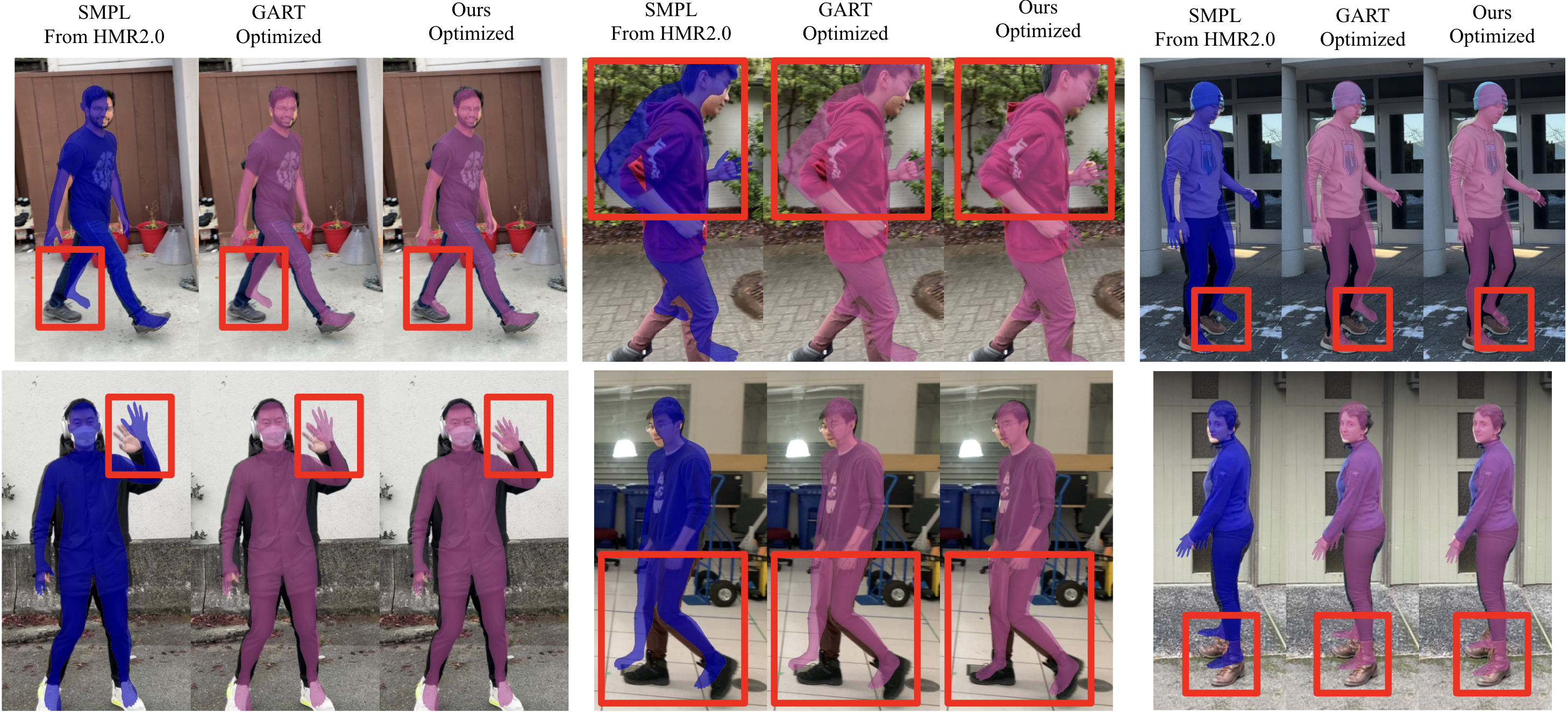}
        \caption{
        SMPL estimation comparison on NeuMan dataset. HumanSplatHMR shows better SMPL refinement results than GART. 
         }
        \label{fig:SMPL_neuman}
\end{figure*}

\subsection{SMPL Evaluation}
We report mean per-joint position error (MPJPE)~\cite{h36m} and Procrustes-aligned MPJPE (PA-MPJPE)~\cite{hmr} on Human3.6M (H36M) and 3DPW in Table~\ref{tab:smpl_pose}.
Additionally, NeuMan results are reported in Table~\ref{tab:neuman-smpl}. H36M is a relatively simple dataset, so HumanSplatHMR only attains slightly improved joint accuracy. In contrast, on 3DPW, our joint mesh–pose refinement consistently yields the lowest error, indicating larger gains when initial poses are noisier, which demonstrates our capability to handle noisier initial SMPL estimation. On the NeuMan dataset, we also demonstrate overall better performance compared to other baselines, as shown in Figure~\ref{fig:SMPL_neuman}.


\begin{figure*}[t]
    \centering
    \includegraphics[width=0.77\linewidth]{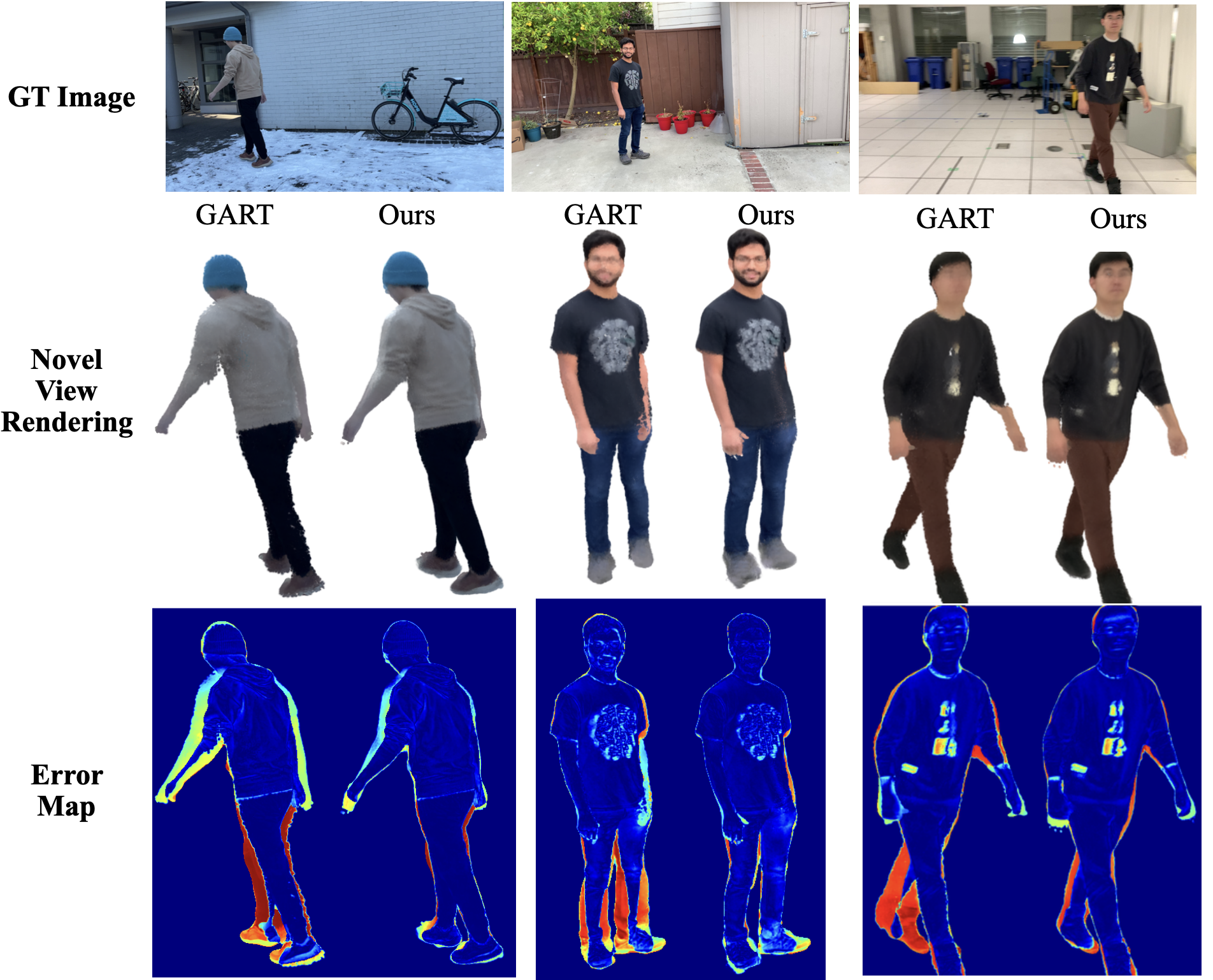}
        \caption{
        Novel view rendering evaluation. HumanSplatHMR shows better rendering quality compared to GART~\cite{gart}. While GART enables learnable skinning weights to handle SMPL pose error from HMR2.0~\cite{4dhuman} in training views, the novel view/pose rendering can still be bad without correctly refining SMPL poses. In contrast, the proposed CAMEL enables joint optimization of both SMPL pose and Gaussian splats, thus outperforming previous works.
        }
        \label{fig:render_neuman}
\end{figure*}
\subsection{Rendering Evaluation}
We evaluate rendering quality on six challenging in-the-wild sequences from the NeuMan dataset under three training setups to cover different train/test ratios and view differences. First, we train \emph{HumanSplatHMR} on 80\% of the frames and evaluate on the held-out 20\% by uniformly withholding every fifth frame. Next, we train on the first 80\% of the frames and test on the final 20\%. Finally, we uniformly sample only 20 camera views for training and use all remaining views for testing. 
Novel view synthesis results for these different train/test schemes are reported in Table~\ref{tab:rendering_neuman}. In all setups, \emph{HumanSplatHMR} consistently outperforms GART. Qualitative results are shown in Figure~\ref{fig:render_neuman}.

\subsection{Ablation Studies}
The ablation study in Table~\ref{tab:ablation-metrics} shows that the full method achieves the best balance of rendering quality and SMPL accuracy. Disabling SMPL optimization or depth supervision leads to noticeable performance degradation. Notably, enabling learnable skinning weights produces rendering quality comparable to the full method but reduces SMPL accuracy, since Gaussians can detach from the mesh and overfit the training views without improving pose estimation, as illustrated in Figure~\ref{fig:learnabe skinning weight}.

We also perform ablation studies on different strategies for binding Gaussians to SMPL. Without any constraint between Gaussians and the SMPL mesh, both rendering quality and SMPL accuracy degrade significantly, as expected. Using learnable skinning weights (LSW), following most existing works~\cite{hugs, gart, gauhuman, splattingavatar, gomavatar}, yields the best rendering quality but poorer SMPL estimation due to the weak connection between Gaussians and the mesh. In contrast, enabling Gaussian-on-Mesh (GoM) embedding produces more accurate human geometry but suboptimal rendering quality, consistent with prior findings~\cite{ihuman}. Finally, our proposed CAMEL achieves the best overall trade-off, delivering strong rendering quality and accurate SMPL estimation by loosely coupling Gaussians with the SMPL mesh.


\begin{table}[t]
\centering
\renewcommand{\arraystretch}{1.1} 
\setlength{\tabcolsep}{2.5pt}
\footnotesize 
\begin{tabular}{l|ccc|ccc}
\toprule
& \multicolumn{3}{c}{Rendering Quality} & \multicolumn{3}{|c}{SMPL Estimation} \\
\cmidrule(lr){2-4}\cmidrule(lr){5-7}
Ablation & psnr$\uparrow$ & ssim$\uparrow$ & lpips$\downarrow$ & mpjpe$\downarrow$ & pa-mpjpe$\downarrow$ & v2v$\downarrow$ \\
\midrule
w/o SMPL Opt.  &  23.52 &  0.960 &  0.0400 &  93.11 &  59.59 &  76.25 \\
w/o Depth Loss  &  25.12 &  0.966 &  0.0337 &  85.58 &  56.79 &  70.49 \\
w/ LSW     &  26.02 &  \textbf{0.970} &  0.0309 &  87.36 &  57.28 &  68.46 \\
Full (Ours)            &  \textbf{26.24} &  \textbf{0.970} &  \textbf{0.0306} &  \textbf{82.88} &  \textbf{56.31} &  \textbf{66.06} \\
\midrule
\multicolumn{6}{l}{Ablation on Gaussian-to-SMPL Binding}\\   
\midrule
No Constraint              &  25.67 &  0.967 &  0.0314 &  84.88 &  57.45 &  67.59 \\
LSW  & \underline{26.16} &\underline{0.968} &\textbf{0.0303} &86.80 &58.29 &69.15\\
GoM             &26.05 &0.968 &0.0314 &\textbf{81.83} & \underline{57.20} &\textbf{65.69}\\
CAMEL            &  \textbf{26.24} &  \textbf{0.970} &  \underline{0.0306} &  \underline{82.88} &  \textbf{56.31} &  \underline{66.06} \\
\bottomrule
\end{tabular}
\caption{Ablation of key components. Left: novel view rendering quality of 3D Gaussian splats. Right: SMPL reconstruction accuracy. We evaluate novel-view/pose synthesis and SMPL accuracy by disabling key components. Results are averaged on $6$ subjects of NeuMan. 'LSW' denotes Learnable Skinning Weights}
\label{tab:ablation-metrics}
\end{table}



\section{Conclusion}
We introduced HumanSplatHMR, a framework that enables Gaussian–SMPL pose joint optimization through our proposed Cloth-Aware Mesh-Embedded Loss (CAMEL). Unlike prior works that face a trade-off between novel-view rendering and mesh reconstruction accuracy, CAMEL supports joint optimization and achieves improvements in both photorealistic human avatar rendering and mesh recovery. Extensive experiments on three public datasets demonstrate consistent gains over three mesh recovery and three human avatar baselines. Our pipeline is flexible and can be integrated with different end-to-end mesh recovery models to further refine SMPL estimation. While our current method focuses on single-human reconstruction, future work will extend HumanSplatHMR to handle multi-human tracking and reconstruction.
\FloatBarrier
{
    \small
    \bibliographystyle{ieeenat_fullname}
    \bibliography{main}
}

\end{document}

%% file: preamble.tex
\usepackage{array}

\usepackage{placeins}


%% file: tables/tab3_horizontal.tex
\begin{table*}[t]
\centering
\footnotesize
\setlength{\tabcolsep}{1.55pt}
\scalebox{0.9}{%
\begin{tabular}{l|ccc|ccc|ccc|ccc|ccc|ccc}
\toprule
\multirow{2}{*}{Method} &
\multicolumn{3}{c|}{Bike} &
\multicolumn{3}{c|}{Citron} &
\multicolumn{3}{c}{Jogging} &
\multicolumn{3}{c|}{Lab} &
\multicolumn{3}{c|}{ParkingLot} &
\multicolumn{3}{c}{Seattle} \\
\cmidrule(lr){2-4}\cmidrule(lr){5-7}\cmidrule(lr){8-10}\cmidrule(lr){11-13}\cmidrule(lr){14-16}\cmidrule(lr){17-19}
 & mpjpe$\downarrow$ & \shortstack{pa-mpjpe}$\downarrow$ & v2v$\downarrow$
 & mpjpe$\downarrow$ & \shortstack{pa-mpjpe}$\downarrow$ & v2v$\downarrow$
 & mpjpe$\downarrow$ & \shortstack{pa-mpjpe}$\downarrow$ & v2v$\downarrow$
 & mpjpe$\downarrow$ & \shortstack{pa-mpjpe}$\downarrow$ & v2v$\downarrow$
 & mpjpe$\downarrow$ & \shortstack{pa-mpjpe}$\downarrow$ & v2v$\downarrow$
 & mpjpe$\downarrow$ & \shortstack{pa-mpjpe}$\downarrow$ & v2v$\downarrow$ \\
\midrule
HMR2.0 & 77.21 & 58.23 & 70.82 & 89.29 & 52.81 & 77.37 & 99.36 & 69.90 & 90.79 & 94.83 & 58.08 & 75.42 & \textbf{69.90} & 51.33 & 56.87 & 95.79 & 55.73 & 74.80 \\
GART & 76.12 & 57.52 & 70.57 & \textbf{87.59} & \textbf{51.96} & 74.33 & 98.25 & 69.50 & 89.82 & 93.57 & 57.51 & 73.07 & 69.95 & \textbf{51.25} & \textbf{55.72} & 93.36 & 54.91 & 72.65 \\
Ours & \textbf{73.62} & \textbf{50.49} & \textbf{56.75} 
     & 88.44 & 52.19 & \textbf{66.54} 
     & \textbf{93.25} & \textbf{67.25} & \textbf{76.83} 
     & \textbf{89.36} & \textbf{55.78} & \textbf{67.11} 
     & 74.73 & 53.77 & 58.97 
     & \textbf{84.55} & \textbf{51.53} & \textbf{67.02} \\
\bottomrule
\end{tabular}
}
\caption{
SMPL accuracy evaluation on NeuMan dataset. Our method achieves overall the best SMPL estimation with lowest error. In this experiment, we deploy uniform 80\% train-split strategy, witholding every fifth frame.
}
\label{tab:neuman-smpl}
\end{table*}

%% file: tables/all_rendering.tex
\begin{table*}[t]
\centering
\small
\setlength{\tabcolsep}{1.55pt} 
\begin{tabular}{c|r|ccc|ccc|ccc|ccc|ccc|ccc}
\toprule
 &  &
\multicolumn{3}{c|}{Bike} &
\multicolumn{3}{c|}{Citron} &
\multicolumn{3}{c|}{Jogging} &
\multicolumn{3}{c|}{Lab} &
\multicolumn{3}{c|}{Parkinglot} &
\multicolumn{3}{c}{Seattle}\\
\cmidrule(lr){3-5}\cmidrule(lr){6-8}\cmidrule(lr){9-11}\cmidrule(lr){12-14}\cmidrule(lr){15-17}\cmidrule(lr){18-20}
&  Method &
psnr$\uparrow$ & ssim$\uparrow$ & lpips$\downarrow$
& psnr$\uparrow$ & ssim$\uparrow$ & lpips$\downarrow$
& psnr$\uparrow$ & ssim$\uparrow$ & lpips$\downarrow$
& psnr$\uparrow$ & ssim$\uparrow$ & lpips$\downarrow$
& psnr$\uparrow$ & ssim$\uparrow$ & lpips$\downarrow$
& psnr$\uparrow$ & ssim$\uparrow$ & lpips$\downarrow$\\
\midrule
\multirow{2}{*}{\shortstack{Uniform\\80\%}} 
  & GART  &23.57 &0.958 &0.044 &24.76 &0.964 &0.031 &24.76 &0.963 &0.038 &23.16 &0.963 &0.041 &27.60 &0.974 &0.031 &27.94 &0.975 &0.021 \\
  & Ours  &\textbf{25.53} &\textbf{0.964} &\textbf{0.033} &\textbf{27.18} &\textbf{0.971} &\textbf{0.022} &\textbf{26.58} &\textbf{0.968} &\textbf{0.029} &\textbf{25.27} &\textbf{0.971} &\textbf{0.030} &\textbf{28.54} &\textbf{0.976} &\textbf{0.024} &\textbf{29.85} &\textbf{0.980} &\textbf{0.016} \\
\midrule
\multirow{2}{*}{\shortstack{20\\Views}} 
  & GART  &22.89 &0.954 &0.045 &24.31 &0.960 &0.033 &24.21 &0.959 &0.040 &22.29 &0.959 &0.044 &26.29 &0.966 &0.037 &27.43 &0.973 &0.022 \\
  & Ours  &\textbf{23.98} &\textbf{0.955} &\textbf{0.036} &\textbf{26.10} &\textbf{0.965} &\textbf{0.025} &\textbf{25.10} &\textbf{0.961} &\textbf{0.033} &\textbf{23.87} &\textbf{0.964} &\textbf{0.034} &\textbf{27.11} &\textbf{0.970} &\textbf{0.029} &\textbf{29.17} &\textbf{0.978} &\textbf{0.017} \\
\midrule
\multirow{2}{*}{\shortstack{First\\80\%}} 
  & GART  &20.97 &0.958 &0.047 &23.28 &0.952 &0.039 &24.40 &0.963 &0.036 &21.67 &0.955 &0.051 &24.06 &0.955 &0.047 &27.01 &0.971 &0.024 \\
  & Ours  &\textbf{24.22} &\textbf{0.965} &\textbf{0.038} &\textbf{24.53} &\textbf{0.959} &\textbf{0.033} &\textbf{25.79} &\textbf{0.965} &\textbf{0.032} &\textbf{24.04} &\textbf{0.963} &\textbf{0.037} &\textbf{25.22} &\textbf{0.959} &\textbf{0.037} &\textbf{28.84} &\textbf{0.976} &\textbf{0.018} \\
\bottomrule
\end{tabular}
\caption{Novel view rendering quality using multiple train-test splits. `Uniform 80\%' indicates witholding every fifth frame, as is standard in the baselines. `20 Views' indicates training on 20 randomly-sampled views. The `First 80\%' split witholds the final 20\% of frames from training, resulting in train and test views that are more dissimilar than alternate train-test schemes, showing the benefits of HumanSplat for rendering configurations far from the training frames.}
\label{tab:rendering_neuman}
\end{table*}

%% file: main.bib
@String(ECCV= {Eur. Conf. Comput. Vis.})

@String(TOG= {ACM Trans. Graph.})

@String(ECCV  = {ECCV})

@String(TOG   = {ACM TOG})

@inproceedings{3dpw,
  title={Recovering accurate 3d human pose in the wild using imus and a moving camera},
  author={Von Marcard, Timo and Henschel, Roberto and Black, Michael J and Rosenhahn, Bodo and Pons-Moll, Gerard},
  booktitle={Proceedings of the European conference on computer vision (ECCV)},
  pages={601--617},
  year={2018}
}

@inproceedings{neuman,
  title={Neuman: Neural human radiance field from a single video},
  author={Jiang, Wei and Yi, Kwang Moo and Samei, Golnoosh and Tuzel, Oncel and Ranjan, Anurag},
  booktitle={European Conference on Computer Vision},
  pages={402--418},
  year={2022},
  organization={Springer}
}

@incollection{huber,
  title={Robust estimation of a location parameter},
  author={Huber, Peter J},
  booktitle={Breakthroughs in statistics: Methodology and distribution},
  pages={492--518},
  year={1992},
  publisher={Springer}
}

@article{lbs,
  title={Fast automatic skinning transformations},
  author={Jacobson, Alec and Baran, Ilya and Kavan, Ladislav and Popovi{\'c}, Jovan and Sorkine, Olga},
  journal={ACM Transactions on Graphics (ToG)},
  volume={31},
  number={4},
  pages={1--10},
  year={2012},
  publisher={ACM New York, NY, USA}
}

@article{h36m,
  title={Human3. 6m: Large scale datasets and predictive methods for 3d human sensing in natural environments},
  author={Ionescu, Catalin and Papava, Dragos and Olaru, Vlad and Sminchisescu, Cristian},
  journal={IEEE transactions on pattern analysis and machine intelligence},
  volume={36},
  number={7},
  pages={1325--1339},
  year={2013},
  publisher={IEEE}
}

@article{unidepthv2,
  title={Unidepthv2: Universal monocular metric depth estimation made simpler},
  author={Piccinelli, Luigi and Sakaridis, Christos and Yang, Yung-Hsu and Segu, Mattia and Li, Siyuan and Abbeloos, Wim and Van Gool, Luc},
  journal={arXiv preprint arXiv:2502.20110},
  year={2025}
}

@inproceedings{sam,
  title={Segment anything},
  author={Kirillov, Alexander and Mintun, Eric and Ravi, Nikhila and Mao, Hanzi and Rolland, Chloe and Gustafson, Laura and Xiao, Tete and Whitehead, Spencer and Berg, Alexander C and Lo, Wan-Yen and others},
  booktitle={Proceedings of the IEEE/CVF international conference on computer vision},
  pages={4015--4026},
  year={2023}
}

@article{film,
  title={Film and television animation production technology based on expression transfer and virtual digital human},
  author={Zhang, Ning and Pu, Belei},
  journal={Scalable Computing: Practice and Experience},
  volume={25},
  number={6},
  pages={5560--5567},
  year={2024}
}

@article{nerf,
  title={Nerf: Representing scenes as neural radiance fields for view synthesis},
  author={Mildenhall, Ben and Srinivasan, Pratul P and Tancik, Matthew and Barron, Jonathan T and Ramamoorthi, Ravi and Ng, Ren},
  journal={Communications of the ACM},
  volume={65},
  number={1},
  pages={99--106},
  year={2021},
  publisher={ACM New York, NY, USA}
}

@article{samv2,
  title={Sam 2: Segment anything in images and videos},
  author={Ravi, Nikhila and Gabeur, Valentin and Hu, Yuan-Ting and Hu, Ronghang and Ryali, Chaitanya and Ma, Tengyu and Khedr, Haitham and R{\"a}dle, Roman and Rolland, Chloe and Gustafson, Laura and others},
  journal={arXiv preprint arXiv:2408.00714},
  year={2024}
}

@article{3dgs,
  title={3D Gaussian splatting for real-time radiance field rendering.},
  author={Kerbl, Bernhard and Kopanas, Georgios and Leimk{\"u}hler, Thomas and Drettakis, George},
  journal={ACM Trans. Graph.},
  volume={42},
  number={4},
  pages={139--1},
  year={2023}
}

@inproceedings{2dgs,
  title={2d gaussian splatting for geometrically accurate radiance fields},
  author={Huang, Binbin and Yu, Zehao and Chen, Anpei and Geiger, Andreas and Gao, Shenghua},
  booktitle={ACM SIGGRAPH 2024 conference papers},
  pages={1--11},
  year={2024}
}

@inproceedings{smpl,
  title={Keep it SMPL: Automatic estimation of 3D human pose and shape from a single image},
  author={Bogo, Federica and Kanazawa, Angjoo and Lassner, Christoph and Gehler, Peter and Romero, Javier and Black, Michael J},
  booktitle={European conference on computer vision},
  pages={561--578},
  year={2016},
  organization={Springer}
}

@article{SMPLify,
  title={Learnable SMPLify: A Neural Solution for Optimization-Free Human Pose Inverse Kinematics},
  author={Yang, Yuchen and Dong, Linfeng and Wang, Wei and Zhong, Zhihang and Sun, Xiao},
  journal={arXiv preprint arXiv:2508.13562},
  year={2025}
}

@inproceedings{hmr,
  title={End-to-end recovery of human shape and pose},
  author={Kanazawa, Angjoo and Black, Michael J and Jacobs, David W and Malik, Jitendra},
  booktitle={Proceedings of the IEEE conference on computer vision and pattern recognition},
  pages={7122--7131},
  year={2018}
}

@inproceedings{phalp,
  title={Decoupling human and camera motion from videos in the wild},
  author={Ye, Vickie and Pavlakos, Georgios and Malik, Jitendra and Kanazawa, Angjoo},
  booktitle={Proceedings of the IEEE/CVF conference on computer vision and pattern recognition},
  pages={21222--21232},
  year={2023}
}

@inproceedings{4dhuman,
  title={Humans in 4d: Reconstructing and tracking humans with transformers},
  author={Goel, Shubham and Pavlakos, Georgios and Rajasegaran, Jathushan and Kanazawa, Angjoo and Malik, Jitendra},
  booktitle={Proceedings of the IEEE/CVF International Conference on Computer Vision},
  pages={14783--14794},
  year={2023}
}

@inproceedings{tokenhmr,
  title={Tokenhmr: Advancing human mesh recovery with a tokenized pose representation},
  author={Dwivedi, Sai Kumar and Sun, Yu and Patel, Priyanka and Feng, Yao and Black, Michael J},
  booktitle={Proceedings of the IEEE/CVF conference on computer vision and pattern recognition},
  pages={1323--1333},
  year={2024}
}

@inproceedings{hsmr,
  title={Reconstructing Humans with a Biomechanically Accurate Skeleton},
  author={Xia, Yan and Zhou, Xiaowei and Vouga, Etienne and Huang, Qixing and Pavlakos, Georgios},
  booktitle={Proceedings of the Computer Vision and Pattern Recognition Conference},
  pages={5355--5365},
  year={2025}
}

@inproceedings{Anim-NeRF,
  title={Animatable NeRF Dynamic Detail Enhancement Based on Residual Deformation Field with Progressive Training},
  author={Yang, Menglei and Han, Yuhang and Zhang, Shenhao and Zhang, Xiaohui},
  booktitle={2025 5th International Conference on Computer Graphics, Image and Virtualization (ICCGIV)},
  pages={161--165},
  year={2025},
  organization={IEEE}
}

@inproceedings{humannerf,
  title={Humannerf: Free-viewpoint rendering of moving people from monocular video},
  author={Weng, Chung-Yi and Curless, Brian and Srinivasan, Pratul P and Barron, Jonathan T and Kemelmacher-Shlizerman, Ira},
  booktitle={Proceedings of the IEEE/CVF conference on computer vision and pattern Recognition},
  pages={16210--16220},
  year={2022}
}

@inproceedings{instantavatar,
  title={Instantavatar: Learning avatars from monocular video in 60 seconds},
  author={Jiang, Tianjian and Chen, Xu and Song, Jie and Hilliges, Otmar},
  booktitle={Proceedings of the IEEE/CVF Conference on Computer Vision and Pattern Recognition},
  pages={16922--16932},
  year={2023}
}

@inproceedings{neuralbody,
  title={Neural body: Implicit neural representations with structured latent codes for novel view synthesis of dynamic humans},
  author={Peng, Sida and Zhang, Yuanqing and Xu, Yinghao and Wang, Qianqian and Shuai, Qing and Bao, Hujun and Zhou, Xiaowei},
  booktitle={Proceedings of the IEEE/CVF conference on computer vision and pattern recognition},
  pages={9054--9063},
  year={2021}
}

@inproceedings{hugs,
  title={Hugs: Human gaussian splats},
  author={Kocabas, Muhammed and Chang, Jen-Hao Rick and Gabriel, James and Tuzel, Oncel and Ranjan, Anurag},
  booktitle={Proceedings of the IEEE/CVF conference on computer vision and pattern recognition},
  pages={505--515},
  year={2024}
}

@inproceedings{gart,
  title={Gart: Gaussian articulated template models},
  author={Lei, Jiahui and Wang, Yufu and Pavlakos, Georgios and Liu, Lingjie and Daniilidis, Kostas},
  booktitle={Proceedings of the IEEE/CVF conference on computer vision and pattern recognition},
  pages={19876--19887},
  year={2024}
}

@inproceedings{gauhuman,
  title={Gauhuman: Articulated gaussian splatting from monocular human videos},
  author={Hu, Shoukang and Hu, Tao and Liu, Ziwei},
  booktitle={Proceedings of the IEEE/CVF conference on computer vision and pattern recognition},
  pages={20418--20431},
  year={2024}
}

@inproceedings{splattingavatar,
  title={Splattingavatar: Realistic real-time human avatars with mesh-embedded gaussian splatting},
  author={Shao, Zhijing and Wang, Zhaolong and Li, Zhuang and Wang, Duotun and Lin, Xiangru and Zhang, Yu and Fan, Mingming and Wang, Zeyu},
  booktitle={Proceedings of the IEEE/CVF Conference on Computer Vision and Pattern Recognition},
  pages={1606--1616},
  year={2024}
}

@inproceedings{gomavatar,
  title={Gomavatar: Efficient animatable human modeling from monocular video using gaussians-on-mesh},
  author={Wen, Jing and Zhao, Xiaoming and Ren, Zhongzheng and Schwing, Alexander G and Wang, Shenlong},
  booktitle={Proceedings of the IEEE/CVF Conference on Computer Vision and Pattern Recognition},
  pages={2059--2069},
  year={2024}
}

@inproceedings{ihuman,
  title={ihuman: Instant animatable digital humans from monocular videos},
  author={Paudel, Pramish and Khanal, Anubhav and Paudel, Danda Pani and Tandukar, Jyoti and Chhatkuli, Ajad},
  booktitle={European Conference on Computer Vision},
  pages={304--323},
  year={2024},
  organization={Springer}
}

@inproceedings{spin,
  title={Learning to reconstruct 3D human pose and shape via model-fitting in the loop},
  author={Kolotouros, Nikos and Pavlakos, Georgios and Black, Michael J and Daniilidis, Kostas},
  booktitle={Proceedings of the IEEE/CVF international conference on computer vision},
  pages={2252--2261},
  year={2019}
}

@inproceedings{hmmr,
  title={Learning 3d human dynamics from video},
  author={Kanazawa, Angjoo and Zhang, Jason Y and Felsen, Panna and Malik, Jitendra},
  booktitle={Proceedings of the IEEE/CVF conference on computer vision and pattern recognition},
  pages={5614--5623},
  year={2019}
}

@inproceedings{vibe,
  title={Vibe: Video inference for human body pose and shape estimation},
  author={Kocabas, Muhammed and Athanasiou, Nikos and Black, Michael J},
  booktitle={Proceedings of the IEEE/CVF conference on computer vision and pattern recognition},
  pages={5253--5263},
  year={2020}
}

@inproceedings{meva,
  title={Meva: A large-scale multiview, multimodal video dataset for activity detection},
  author={Corona, Kellie and Osterdahl, Katie and Collins, Roderic and Hoogs, Anthony},
  booktitle={Proceedings of the IEEE/CVF winter conference on applications of computer vision},
  pages={1060--1068},
  year={2021}
}

@inproceedings{sadgs,
  title={Sad-gs: Shape-aligned depth-supervised gaussian splatting},
  author={Kung, Pou-Chun and Isaacson, Seth and Vasudevan, Ram and Skinner, Katherine A},
  booktitle={Proceedings of the IEEE/CVF Conference on Computer Vision and Pattern Recognition},
  pages={2842--2851},
  year={2024}
}

@inproceedings{autosplat,
  title={Autosplat: Constrained gaussian splatting for autonomous driving scene reconstruction},
  author={Khan, Mustafa and Fazlali, Hamidreza and Sharma, Dhruv and Cao, Tongtong and Bai, Dongfeng and Ren, Yuan and Liu, Bingbing},
  booktitle={2025 IEEE International Conference on Robotics and Automation (ICRA)},
  pages={8315--8321},
  year={2025},
  organization={IEEE}
}

@inproceedings{splatface,
  title={SplatFace: Gaussian splat face reconstruction leveraging an optimizable surface},
  author={Luo, Jiahao and Liu, Jing and Davis, James},
  booktitle={2025 IEEE/CVF Winter Conference on Applications of Computer Vision (WACV)},
  pages={774--783},
  year={2025},
  organization={IEEE}
}

@inproceedings{sugar,
  title={Sugar: Surface-aligned gaussian splatting for efficient 3d mesh reconstruction and high-quality mesh rendering},
  author={Gu{\'e}don, Antoine and Lepetit, Vincent},
  booktitle={Proceedings of the IEEE/CVF Conference on Computer Vision and Pattern Recognition},
  pages={5354--5363},
  year={2024}
}

@ARTICLE{ssim,
  author={Zhou Wang and Bovik, A.C. and Sheikh, H.R. and Simoncelli, E.P.},
  journal={IEEE Transactions on Image Processing}, 
  title={Image quality assessment: from error visibility to structural similarity}, 
  year={2004},
  volume={13},
  number={4},
  pages={600-612},
  doi={10.1109/TIP.2003.819861}}
